\documentclass[runningheads]{llncs}
\usepackage{graphicx}
\usepackage{amsmath}

\begin{document}
\title{Combining Heterogeneously Labeled Datasets For Training Segmentation Networks}
\titlerunning{Combining Heterogeneously Labeled Datasets in Segmentation Networks}
% If the paper title is too long for the running head, you can set
% an abbreviated paper title here
%
\author{Jana Kemnitz\inst{1,2,3} \and
Christian F. Baumgartner\inst{3} \and
Wolfgang Wirth\inst{1,2} \and
Felix Eckstein\inst{1,2} \and
Sebastian K. Eder\inst{1} \and
Ender Konukoglu\inst{3}}

\authorrunning{J. Kemnitz et al.}
% First names are abbreviated in the running head.
% If there are more than two authors, 'et al.' is used.
%

%
\institute{
Paracelsus Medical University Salzburg, Austria\and
Chondrometrics GmbH Ainring, Germany
\and
Computer Vision Lab, ETH Zurich, Switzerland\\
}

%\author{*****************\inst{*} \and
%*************\inst{*} \and
%***************\inst{*} \and
%*************\inst{*} \and
%***********\inst{*} \and
%*************\inst{*}}
%
%\authorrunning{****** et al.}
% First names are abbreviated in the running head.
% If there are more than two authors, 'et al.' is used.
%

%
%\institute{
%**\and
%***
%\and
%*****\\
%}

%
\maketitle              % typeset the header of the contribution
\begin{abstract}
Accurate segmentation of medical images is an important step towards analyzing and tracking disease related morphological alterations in the anatomy. Convolutional neural networks (CNNs) have recently emerged as a powerful tool for many segmentation tasks in medical imaging. The performance of CNNs strongly depends on the size of the training data and combining data from different sources is an effective strategy for obtaining larger training datasets. However, this is often challenged by heterogeneous labeling of the datasets. For instance, one of the dataset may be missing labels or a number of labels may have been combined into a super label. In this work we propose a cost function which allows integration of multiple datasets with heterogeneous label subsets into a joint training. We evaluated the performance of this strategy on thigh MR and a cardiac MR datasets in which we artificially merged labels for half of the data. We found the proposed cost function substantially outperforms a naive masking approach, obtaining results very close to using the full annotations. 

%WHAT ARE YOU COMPARING TO IN THOSE NUMBERS? I.E. WHAT IS THE SECOND NUMBER? We found that combining the two data sets led to better segmentation performance in thigh (DSC: 0.940 (0.054) vs 0.847 (0.124)) and cardiac (DSC: 0.901 (0.030) vs 0.889 (0.036)) MRIs.  

% \keywords{Deep Learning\and Training\and Combining Data\and Thigh\and Cardiac\and MRI}  % We can get rid of this to safe space
\end{abstract}
\section{Introduction}
Accurate segmentation of complex and anatomical structures in medical images is the one of most critical parts in the image analysis pipeline. Segmentation results affect all the subsequent processes of image analysis such as object representation, feature measurement, the development of imaging biomarkers and ultimately the resulting diagnosis and treatment of diseases~\cite{ref_shen2017,ref_Prescott2013}.

The recent reemergence of convolutional neural networks (CNNs) allows automatic segmentation of anatomical structures with unprecedented accuracy~\cite{ref_litjens2017,ref_ronneberger2015}. However, the performance of CNNs depends strongly on the size of the training data~\cite{ref_litjens2017}. Since fully annotated datasets are still often relatively small, a possible strategy is to combine multiple datasets from different sources for training. 

Apart from possible domain shifts, a problem that may arise in practice is that different datasets may be following different labeling protocols and may thus contain different subsets of labels. For instance, detailed labels in one dataset may be combined into a ``super label'' in another dataset, or a label may be completely missing from the one of the datasets. Note that the latter case can be thought of as the missing label forming a super label with the background. 

Combining heterogeneously labeled datasets has previously been investigated in the context of atlas-based segmentation employing majority voting, semilocally weighted voting, performance level estimation and multi-protocol label fusion~\cite{Iglesias_2014}. However, to our knowledge incorporating such data for training segmentation networks still remains a open challenge~\cite{ref_litjens2017}.
%Combining heterogeneously labeled datasets has already been critical for previous segmentation approaches as atlas-based segmentation~\cite{Iglesias_2014} while to our knowledge incorporating this uncertainty directly in the cost function of segmentation networks still remains an open challenge~\cite{ref_litjens2017}. 

A naive approach to address this problem would be to simply set training cost function (e.g. crossentropy loss) to zero at pixel locations where the desired label is not available. This means that in those locations the network would be free to predict any label. However, this is not taking full advantage of the available information. For instance, for training images for which one label is missing, we know that in those locations the network should only predict background or the missing label, but not any other labels. Similarly, if a training image combines two anatomic labels into one, in those regions only those two structures should be predicted, but not, for example, background.

%One specific motivation of this work is to investigate the potential of a cost function combining relatively small completely~\cite{ref_ruhdorfer2013,ref_ruhdorfer2015} and relatively large incompletely \cite{ref_kemnitz2018,Culvenor_2018} annotated thigh MR label subsets available from previous studies from the Osteoarthritis Initiative database (OAI)~\cite{ref_nevitt2006} in a joint network training.

In this paper, we propose a simple and effective cost function which allows integrating such information into the training process and thus takes advantage of the full extent of available training information. We evaluate the proposed cost function on two datasets: thigh MR images from the Osteoarthritis Initiative (OAI)~\cite{ref_peterfy2008} and publicly available cardiac MR data from the ACDC challenge~\cite{ref_Bernard_2018}. For both datasets we simulate incomplete labels by merging a number of labels for parts of the datasets.

\section{Methods} \label{sec:methods}

The goal of the proposed method is to learn the parameters of a segmentation network which can assign a label $y_i \in \mathcal{L}_a = \{\ell_0,\ldots,\ell_L\}$ for each pixel $i$ of an image $X$. Generally, for training we may have multiple datasets which have been annotated with different subsets of those labels. To describe the proposed method, we focus on the simpler problem, where we assume that we have only two training datasets $D_1,D_2$ of which $D_1$ was annotated with all target labels, while $D_2$ contains one super label $\mathcal{L}_s = \{\ell_0, \ldots, \ell_S\}$ that corresponds to $S$ of the labels in $\mathcal{L}_a$. That is $D_2$ contains the following labels $\{ \{\ell_0, \ldots, \ell_S \}, \ell_{S+1}, \ldots, \ell_L \} = \{ \mathcal{L}_s, \ell_{S+1} \ldots, \ell_L \}$. For notational simplicity we define a binary mask $m_i \in \{0,1\}$ which is 0 at all pixels that have label $\mathcal{L}_s$ and 1 otherwise. In other words, $m_i=1$ only where full information is available. This simplified problem, with two datasets and one super label, can be easily extended to more complex scenarios. 

\begin{figure}
\includegraphics[width=\textwidth]{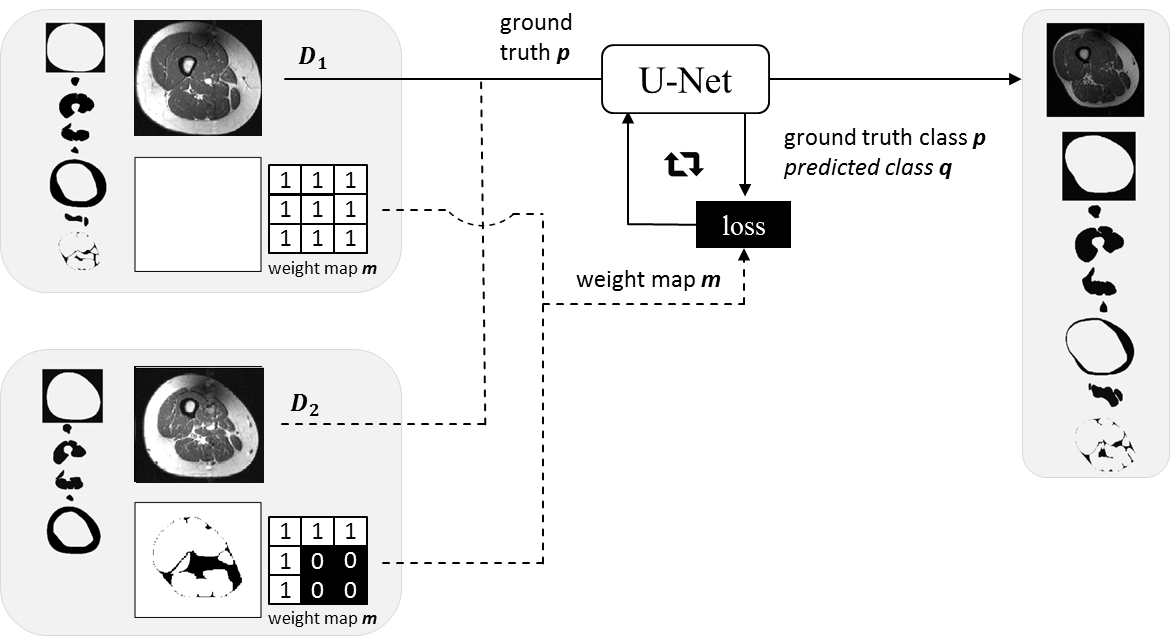}
\caption{Thigh MRI trainings paths of the U-Net for segmentation with a) training data $D_1$ annotated with all target labels, while training dataset $D_2$ contains the following labels $\{ \{\ell_0, \ldots, \ell_S \}, \ell_{S+1}, \ldots, \ell_L \} = \{ \mathcal{L}_s, \ell_{S+1} \ldots, \ell_L \}$; both with binary mask $m_i \in \{0,1\}$ which is 0 at all pixels that have label $\mathcal{L}_s$ and 1 otherwise.} \label{fig1}
\end{figure}

The commonly used cross entropy function for a single fully annotated image is given by
\begin{equation}\label{eq:xent}
\mathcal{C}^{xent} = \sum_i \sum_{\ell \in \mathcal{L}_a} p(y_i=\ell) \log q(y_i=\ell|X),
\end{equation}
where $p$ denotes the ground-truth probability distribution and $q$ denotes the networks softmax output. In the following we consider a naive extension of this cost function disregarding pixels with incomplete information, and our proposed cost function which takes into account the possible predictions of super labels. An overview of the strategies is shown in Fig. \ref{fig1}.

\subsection{Naive Masking}

Apart from completely disregarding datasets with incomplete labeling, the simplest strategy is to mask out regions with incomplete information in the crossentropy loss function:
\begin{equation}\label{eq:naive}
\mathcal{C}^{naive} = \sum_i m_i \sum_{\ell \in \mathcal{L}_a} p(y_i=\ell) \log q(y_i=\ell|X),
\end{equation}
using the mask $m_i$ defined earlier. While still using images from both datasets $D_1, D_2$ for training, this formulation disregards the information contained in $\mathcal{L}_s$, that it corresponds to $\{\ell_0, \ldots, \ell_S \}$ and not to any other label. In practice, we found that this often leads to undesired structure labels or background leaking into those regions. 

\subsection{Super Label Aware Crossentropy Loss}

In order to overcome this limitation, we propose adding an additional term to the crossentropy loss also taking into account the super labels as follows:

\begin{equation}\label{eq:slac}
\begin{split}
\mathcal{C}^{slac}_i & = m_i \sum_{\ell \in \mathcal{L}_a} p(y_i=\ell) \log q(y_i=\ell) + (1 - m_i) \sum_{\ell \in \mathcal{L}_s} p(y_i=\ell) \log \sum_{\ell \in \mathcal{L}_s} q(y_i=\ell), \\
                     & = m_i \sum_{\ell \in \mathcal{L}_a} p(y_i=\ell) \log q(y_i=\ell) + (1 - m_i) \log \sum_{\ell \in \mathcal{L}_s} q(y_i=\ell),
\end{split}
\end{equation}
where we omitted the sum over $i$ and the conditioning on $X$ for brevity. Here, the second term encourages the network to predict $q(y_i=\mathcal{L}_s) = \sum_{\ell \in \mathcal{L}_s} q(y_i=\ell)$, in regions where the training image is labeled with the super label. The simplification in the second equality is due the fact that by definition $\sum_{\ell \in \mathcal{L}_s} p(y_i=\ell)=1$ where $m_i=1$.

% \subsection{Loss Function}\label{sec:loss}
% We trained the U-Net as introduced above from scratch and investigated three different cost functions combining the complete and incomplete training data sets. First, we used the standard pixel-wise cross entropy simply ignoring the incomplete training data set. 

% Subsequently, we combined the complete and incomplete training data sets with naive masking of the unknown areas to take the value of the missing labels and multiplied the pixel-wise cross entropy loss with a pixel wise binary weight mask \(m\in \lbrace 0,1\rbrace ^N\) with 1 indicating that the ground truth for corresponding pixel is available (see Fig.~\ref{fig1}). This means any label can appear, which may cause leaking of other structures in that area.

\section{Experiments and Results}
\subsection{Data} \label{sec:data}

We evaluated segmentation accuracy of the cost functions introduced above on two data sets.
\vskip 0.1in
\textbf{Thigh MRI:} The thigh MRI data consist of 139 patient scans of the osteoarthritis initiative (OAI)~\cite{ref_peterfy2008}, a publicly available data base created for imaging biomarker validation in knee osteoarthritis. MRIs were acquired using a 3T system (slice thickness 5mm; in-plane resolution 0.98mm; no inter-slice gap) and segmentations were available for patient from previous studies~\cite{ref_ruhdorfer2013,ref_ruhdorfer2015}. The dataset was divided into training, test and validation set comprising 99, 20 and 20 subjects, respectively. All muscle MRI slices where cropped and centered towards the femoral bone of the right knee to simplify the segmentation problem with a resulting image size of 256x256 pixels.  %All muscle and cardiac image was intensity normalized to zero mean and unit variance.

%using a 3-T scanner (Siemens Trio, Siemens AG, Erlangen, Germany) and the imaging protocol included T1-weighted spin echo MRIs (slice thickness 5 mm; in-plane resolution 0.98 mm; no inter-slice gap, repetition time 500 ms, echo time 10 ms) of the thigh. All 

\vskip 0.1in
\textbf{Cardiac MRI:} The cardiac MRI data from the ACDC challenge~\cite{ref_Bernard_2018} consists 100 patient scans each including a short-axis cine-MRI acquired on 1.5T and 3T systems with resolutions ranging from 0.70mm to 1.92mm in-plane and 5mm to 10mm through-plane. Segmentation for the background, the myocardium (Myo), the left ventricle (LV) and the right ventricle (RV) were available for the end-diastolic (ED) and end-systolic (ES) phases of each patient. The dataset was divided into training, test and validation set comprising 60, 25 and 15 subjects, respectively. All images were resampled to a common resolution of 1.37x1.37mm$^2$ and resampled centrally placed into images of constant size, padding with zeros where necessary.

\subsection{Network Architecture and Training} 
All experiments were performed using the modified 2D U-Net architecture proposed in~\cite{ref_baumgartner2017}. We used mini-batch gradient descent and the ADAM optimizer with a learning rate of 0.01 to minimize the respective cost functions. The final model was selected based on the respective loss functions evaluated on the validation set.

 % \subsection{Evaluation Measures} 
 % We evaluated the segmentation accuracy achieved with the introduced methods using three measures: the Dice coefficient, the Hausdorff distance and the average symmetric surface distance (ASSD). [To save space this doesn't need to be a Section. We can just mention it somewhere.]

 \subsection{Evaluation} 

% To create a scenario of having complete and incomplete training data available we defined Set I as complete data set and Set II incomplete as described in Table 1 for thigh and cardiac labels. All images were preprocessed as described in Sec. \ref{sec:prepro}.

In order to evaluate the ability of the loss functions discussed in Section \ref{sec:methods} to address the problem of differently labeled datasets, we artificially generated a fully annotated dataset $D_1$ and a dataset $D_2$ for which a number of labels have been merged into super labels. For both the cardiac and thigh datasets we relabeled half of the training and validation sets as summarized in Table \ref{tab:data_labeling}. To generate $D_2$, for the thigh data we merged the AD and IMF labels, and for the cardiac data we created a ``heart'' super label containing all of the structures apart from background. The final performance was evaluated on the fully labeled test sets using the Dice score (DSC), average symmetric surface distance (ASSD) and Hausdorff distance (HD).

\begin{table}[]
\centering
\caption{Simulated data $D_1$ (completely labeled) and $D_2$ (containing a super label $\mathcal{L}_s$) for thigh and cardiac MR segmentation.}
\label{tab:data_labeling}
\begin{tabular}{l|ccllcc}
\textbf{thigh}          & $D_1$ & $D_2$ & \qquad \qquad & \multicolumn{1}{l|}{\textbf{cardiac}}       & $D_1$ & $D_2$ \\ \cline{1-3} \cline{5-7} 
background               & x     & x      &  & \multicolumn{1}{l|}{background}              & x     & x      \\
femoral bone (FB)       & x     & x      &  & \multicolumn{1}{l|}{left ventricular (LV)}  & x     & $\mathcal{L}_s$       \\
quadriceps (QC)         & x     & x      &  & \multicolumn{1}{l|}{right ventricular (RV)} & x     & $\mathcal{L}_s$       \\
flexors (FX)            & x     & x      &  & \multicolumn{1}{l|}{myocardium (Myo)}       & x     & $\mathcal{L}_s$       \\
sartorius (ST)          & x     & x      &  &                                             &       &        \\
subcutaneous fat (SCF)  & x     & x      &  &                                             &       &        \\
adductors (AD)          & x     & $\mathcal{L}_s$       &  &                                             &       &        \\
intermuscular fat (IMF) & x     & $\mathcal{L}_s$       &  &                                             &       &       
\end{tabular}
\end{table}

In addition to the network training with the two cost functions $\mathcal{C}^{naive}$ and $\mathcal{C}^{slac}$ we also evaluated two baseline methods: 1) we trained only on the complete dataset $D_1$ with the normal crossentropy cost function $\mathcal{C}^{xent}$ to obtain a lower bound on the performance, and 2) we trained with $\mathcal{C}^{xent}$ on the entire unaltered training sets to obtain an upper bound.

%  In the first experiment we focused on the U-Net training with all the data using the standard pixel-wise cross entropy loss as described in Sec.\ref{sec:loss} to indicate the upper baseline for our following experiments which can be achieved when all data is complete. 
% Subsequently we created a scenario defining Set I as complete data set and Set II as incomplete data set as described in Sec.\ref{sec:prepro} to simulate the combination of both. In the second experiment we used the standard pixel-wise cross entropy loss in Set I, simply ignoring the incomplete training data Set II, as the easiest approach handling with incomplete data indicating the lower baseline. 
% In the third experiment we combined data Set I and Set II using the naive weight mask masking as described in Sec.\ref{sec:loss}. In the last experiment we combined data Set I and Set II using the modified weight mask masking as described in Sec.\ref{sec:loss}. 
% In Table \ref{table_2} we report the Dice score, ASSD and HD averaged for each label. It can be seen that the small data set and the large data set using a standard pixel-wise cross entropy are indicating the lower and upper baseline as expected. The naive weight mask masking adding Set II leads to a significant improvement and the modified weight mask masking method almost archives the performance of the upper baseline as using the large complete data set (see Fig.~\ref{fig2}).

\begin{table}[]
\centering
\caption{Thigh and cardiac MR segmentation accuracy measure in mean (std) for the evaluated cost functions $\mathcal{C}^{naive}$ and $\mathcal{C}^{slac}$ (best performance in bold font) and the lower bound (LB) and upper bound (UB) for all structures.}
\label{tab:results}
\begin{tabular}{lccccccc}

\multicolumn{8}{c}{\textbf{thigh}}                                                                                             \\ \hline
                & \multicolumn{3}{c}{femoral bone (FB)}       &  & \multicolumn{3}{c}{quadriceps (QC)}         \\ \cline{2-4} \cline{6-8} 
                & DSC           & ASSD        & HD            &  & DSC           & ASSD        & HD            \\ \cline{2-8} 
$\mathcal{C}^{xent}$ (LB) & 0.971 (0.014) & 0.60 (0.77) & 7.00 (12.17) &  & 0.952 (0.056) & 1.32 (0.79) & 14.40 (6.32)  \\
$\mathcal{C}^{naive}$    & \textbf{0.978 (0.008)} & 0.45 (0.58) & 5.40 (12.50) &  & 0.977 (0.006) & 0.81 (0.35) & 10.60 (8.42)   \\
$\mathcal{C}^{slac}$ & 0.974 (0.008) & \textbf{0.38 (0.09)} & \textbf{2.05 (1.70)}  &  & \textbf{0.980 (0.010)} & \textbf{0.61 (0.23)} & \textbf{7.78 (3.79)}   \\
$\mathcal{C}^{xent}$ (UB) & 0.978 (0.006) & 0.32 (0.09) & 1.86 (1.15)  &  & 0.979 (0.008) & 0.67 (0.31) & 7.37 (5.16)   \\ \hline
                & \multicolumn{3}{c}{flexors (FX)}            &  & \multicolumn{3}{c}{sartorius (ST)}          \\ \cline{2-4} \cline{6-8} 
                & DSC           & ASSD        & HD           &  & DSC           & ASSD        & HD            \\ \cline{2-8} 
$\mathcal{C}^{xent}$ (LB)  & 0.905 (0.065) & 2.30 (1.28) & 16.56 (4.88)  &  & 0.809 (0.117) & 4.22 (2.53) & 36.44 (25.82) \\
$\mathcal{C}^{naive}$    & \textbf{0.957 (0.019)} & 1.05 (0.58) & 11.20 (6.51)  &  & 0.903 (0.052) & 1.76 (1.32) & 15.90 (11.07)  \\
$\mathcal{C}^{slac}$ & \textbf{0.957 (0.021)} & \textbf{0.90 (0.35)} & \textbf{9.36 (4.16)}  &  & \textbf{0.967 (0.010)} & \textbf{0.33 (0.09)} & \textbf{2.10 (1.52)}   \\
$\mathcal{C}^{xent}$ (UB)  & 0.968 (0.013) & 0.75 (0.33) & 6.70 (3.71)  &  & 0.945 (0.055) & 0.92 (1.08) & 14.07 (24.17) \\ \hline
                & \multicolumn{3}{c}{subcutanous fat (SCF)}   &  & \multicolumn{3}{c}{adductors (AD)}          \\ \cline{2-4} \cline{6-8} 
                & DSC           & ASSD        & HD           &  & DSC           & ASSD        & HD            \\ \cline{2-8} 
% * <j.kemnitz@outlook.com> 2018-06-17T17:17:41.385Z:
%
% ^.
$\mathcal{C}^{xent}$ (LB)  & 0.936 (0.132) & 0.92 (1.37) & 11.29 (15.28) &  & 0.809 (0.117) & 4.22 (2.53) & 36.44 (25.82) \\
$\mathcal{C}^{naive}$    & 0.965 (0.035) & 0.48 (0.19) & 6.33 (12.38)  &  & 0.908 (0.039) & 1.13 (0.51) & 10.8 (8.66)  \\
$\mathcal{C}^{slac}$ & \textbf{0.974 (0.008)} & \textbf{0.41 (0.09)} & \textbf{6.11 (11.18)}  &  & \textbf{0.967 (0.010)} & \textbf{1.09 (0.45)} & \textbf{9.05 (4.48)}   \\
$\mathcal{C}^{xent}$ (UB)  & 0.975 (0.014) & 0.38 (0.12) & 5.19 (11.93)  &  & 0.945 (0.055) & 0.92 (1.08) & 14.07 (24.17) \\ \hline
                & \multicolumn{3}{c}{intermuscular fat (IMF)} &  &               \multicolumn{3}{c}{\textbf{average}}               \\ \cline{2-4} \cline{6-8}
                & DSC           & ASSD        & HD           &   & DSC           & ASSD        & HD               \\ \cline{2-4} \cline{6-8}
$\mathcal{C}^{xent}$ (LB)  & 0.608 (0.093) & 2.62 (1.09) & 32.67 (10.24) &  & 0.847 (0.100) & 2.05 (1.42) & 18.96 (11.81)               \\
$\mathcal{C}^{naive}$    & 0.744 (0.076) & 1.55 (0.34) & 27.00 (7.31)  &  & 0.919 (0.077) & 1.04 (0.46) & 12.52 (6.70)             \\
$\mathcal{C}^{slac}$& \textbf{0.823 (0.031)} & \textbf{0.92 (0.16)} & \textbf{18.00 (3.29)}  &  & \textbf{0.940 (0.054)} & \textbf{0.66 (0.28)} & \textbf{7.78 (5.20)}               \\
$\mathcal{C}^{xent}$ (UB)   & 0.821 (0.046) & 1.03 (0.36) & 21.96 (7.06)  &  & 0.943 (0.020) & 0.71 (0.31) & 9.29 (7.20)              \\ 
\hline \hline
  \\

\multicolumn{8}{c}{\textbf{cardiac}}                                                                                             \\ \hline
                & \multicolumn{3}{c}{left ventricle (ED)}     &  & \multicolumn{3}{c}{left ventricle (ES)}    \\ \cline{2-4} \cline{6-8} 
                & DSC           & ASSD        & HD            &  & DSC           & ASSD        & HD           \\ \cline{2-8} 
$\mathcal{C}^{xent}$ (LB) & 0.960 (0.018) & 0.37 (0.38) & 5.85 (3.77)   &  & 0.914 (0.040) & 0.81 (0.69) & 8.30 (3.59)  \\
$\mathcal{C}^{naive}$   & 0.951 (0.018) & 0.64 (0.56) & 8.91 (5.78)   &  & 0.919 (0.040) & 1.00 (1.16) & 10.11 (5.69) \\
$\mathcal{C}^{slac}$ & \textbf{0.962 (0.018)} & \textbf{0.42 (0.54)} & \textbf{5.88 (3.64)}   &  & \textbf{0.923 (0.052)} & \textbf{0.77 (0.84)} & \textbf{7.20 (3.16)}  \\
$\mathcal{C}^{xent}$ (UB)  & 0.962 (0.017) & 0.39 (0.48) & 5.49 (2.95)   &  & 0.934 (0.034) & 0.53 (0.40) & 7.76 (3.34)  \\ \hline
                & \multicolumn{3}{c}{right ventricle (ED)}    &  & \multicolumn{3}{c}{right ventricle (ES)}   \\ \cline{2-4} \cline{6-8} 
                & DSC           & ASSD        & HD            &  & DSC           & ASSD        & HD           \\ \cline{2-8} 
$\mathcal{C}^{xent}$ (LB) & 0.876 (0.171) & 1.69 (3.37) & 16.28 (14.57) &  & 0.828 (0.140) & 1.81 (2.26) & 15.96 (7.84) \\
$\mathcal{C}^{naive}$    & 0.909 (0.039) & 0.91 (0.54) & 14.52 (6.78)  &  & 0.809 (0.089) & 2.06 (0.91) & 15.87 (5.51) \\
$\mathcal{C}^{slac}$ & \textbf{0.922 (0.048)} & \textbf{0.83 (0.98)} & \textbf{13.57 (6.13)}  &  & \textbf{0.827 (0.116)} & \textbf{1.76 (1.33)} & \textbf{15.15 (5.96)} \\
$\mathcal{C}^{xent}$ (UB)  & 0.927 (0.043) &0.82 (0.90) & 13.74 (6.33)  &  & 0.834 (0.108)& 1.74 (1.43) & 15.93 (5.73) \\ \hline
                & \multicolumn{3}{c}{myocardium (ED)}         &  & \multicolumn{3}{c}{myocardium (ES)}        \\ \cline{2-4} \cline{6-8} 
                & DSC           & ASSD        & HD            &  & DSC           & ASSD        & HD           \\ \cline{2-8} 
$\mathcal{C}^{xent}$ (LB)  & 0.873 (0.031) & 0.47 (0.18) & 8.17 (5.08)   &  & 0.882 (0.042) & 0.75 (0.47) & 11.80 (5.85) \\
$\mathcal{C}^{naive}$    & 0.852 (0.044) & 0.66 (0.32) & 11.27 (6.58)  &  & 0.863 (0.055) & 0.86 (0.51) & 11.78 (5.85) \\
$\mathcal{C}^{slac}$ & \textbf{0.878 (0.030)} & \textbf{0.54 (0.29)} & \textbf{9.99 (8.46)}   &  & \textbf{0.891 (0.035)} & \textbf{0.67 (0.42)} & \textbf{10.06 (5.68)} \\
$\mathcal{C}^{xent}$ (UB)  & 0.881 (0.026) & 0.51 (0.21) & 8.90 (6.36)   &  & 0.896 (0.039) & 0.61 (0.32) & 10.74 (6.39) \\ \hline

                & \multicolumn{3}{c}{\textbf{average}}     &  &    \\ \cline{2-4} 
                & DSC           & ASSD        & HD            &  &            &        &           \\ \cline{2-4} 
$\mathcal{C}^{xent}$ (LB)  &  0.889 (0.074) & 0.89 (1.32) & 11.06 (6.78)  \\
$\mathcal{C}^{naive}$    &  0.884 (0.048) & 1.02 (1.67) & 12.08 (6.03) \\
$\mathcal{C}^{slac}$ &  \textbf{0.901 (0.050)} & \textbf{0.83 (0.73)} & \textbf{10.31 (5.51)}  \\
$\mathcal{C}^{xent}$ (UB)  &  0.906 (0.045) & 0.77 (0.62) & 10.43 (5.18)  \\ \hline

\end{tabular}
\end{table}

\begin{figure}
\includegraphics[width=\textwidth]{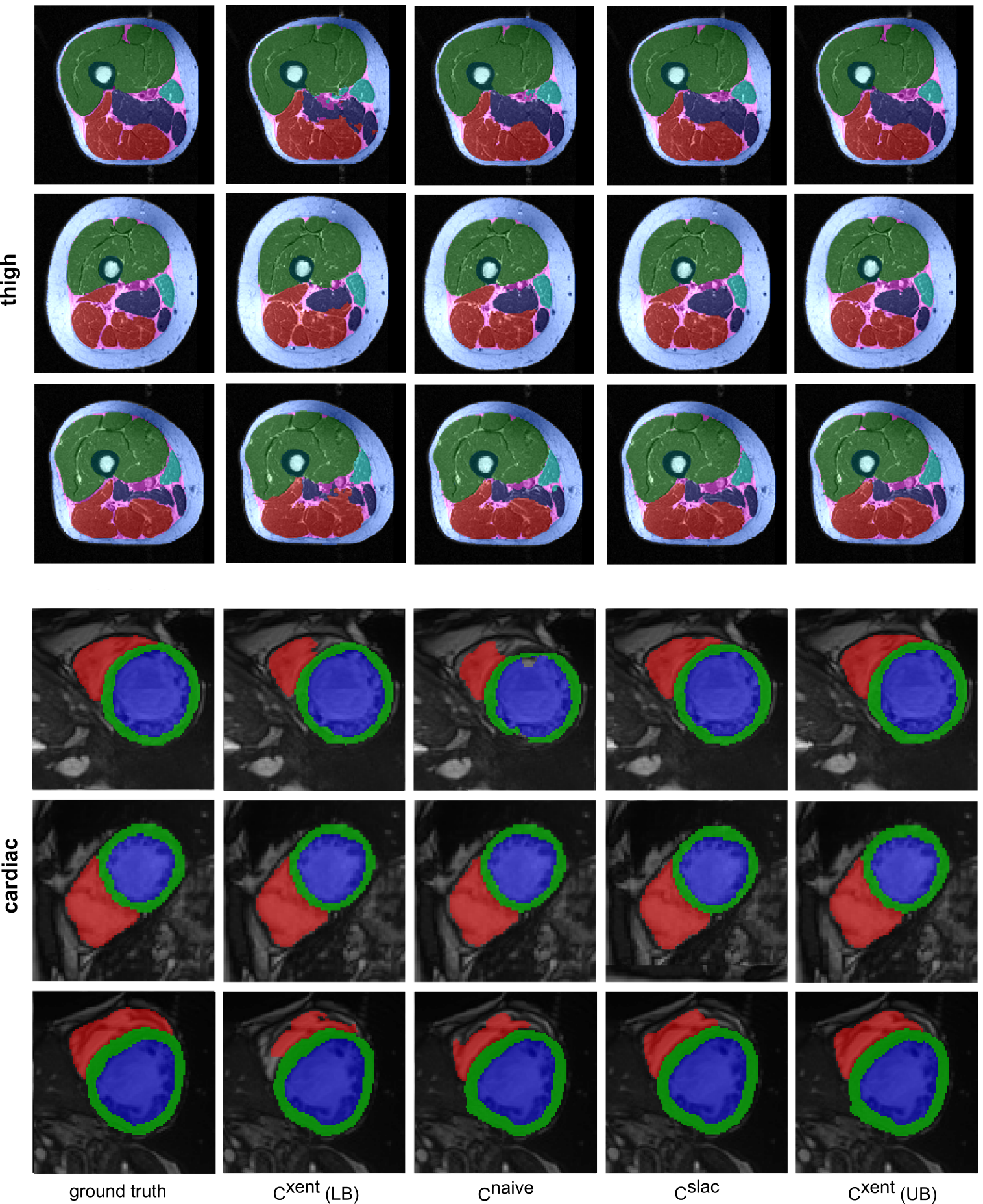}
\center
\caption{Examples of thigh and cardiac ground truth and predicted segmentation using the evaluated cost functions $\mathcal{C}^{naive}$ and $\mathcal{C}^{slac}$ and the lower bound (LB) and upper bound (UB). } \label{fig2}
\end{figure}

The results obtained with the investigated costs functions are summarized in Table \ref{tab:results}. Example segmentations for both datasets are shown in Fig. \ref{fig2}. With the proposed cost function $\mathcal{C}^{slac}$ we achieved segmentation results very close to using the full annotations (upper bound) in both thigh and cardiac datasets.

\subsection{Discussion and Conclusion}

In this work we proposed a cost function to enable the integration of multiple datasets with heterogeneous label subsets into a joint training. We evaluated the performance of this strategy on thigh MR and a cardiac MR datasets in which we artificially merged labels for half of the data. We found the proposed cost function substantially outperforms a naive masking approach and achieved results very close to using the full annotations. This novel cost function improves the segmentation performance compared to a naive masking precisely in those single labeled regions merged into a super label by avoiding undesired label or background leaking (see Fig. \ref{fig2}, Table \ref{tab:results}). 
As expected we found that the proposed cost function led to the biggest improvement over the naive masking approach in regions were labels were merged into super labels. 

One specific motivation of this work was to investigate the potential of this novel loss term in the scope of the OAI database where several datasets with heterogeneous label subsets are available from previous studies \cite{ref_ruhdorfer2013,ref_ruhdorfer2015,ref_kemnitz2018}. This new loss term will allow us to merge all this heterogeneous label subsets into a joint training. 

%This is an important step towards a precise and fully automated segmentation framework to evaluate the thigh morphology from the entire OAI database with regard to various musculoskeletal diseases having over 10.000 MR datasets available. As the amount of publicly available, heterogeneously labeled data is continuously growing, segmentation networks that can integrate multiple datasets with heterogeneous label subsets will become increasingly important. We have shown on thigh MR and cardiac MR data that integrating heterogeneous labeled datasets is feasible with the proposed cost function. Even through, the method was only demonstrated for combining two datasets, we believe that the approach can easily be extended to more complex scenarios.

%
% the environments 'definition', 'lemma', 'proposition', 'corollary',
% 'remark', and 'example' are defined in the LLNCS documentclass as well.
%

%
% ---- Bibliography ----
%
% BibTeX users should specify bibliography style 'splncs04'.
% References will then be sorted and formatted in the correct style.
%
% \bibliographystyle{splncs04}
% \bibliography{mybibliography}
%

% Better use Bibtex!! 

\end{document}